\documentclass[manuscript]{acmart}
\AtBeginDocument{%
  }

\usepackage{multirow}
\usepackage{subcaption} 
\usepackage{algorithmic}
\usepackage{amsmath}
\usepackage{algorithm}
\usepackage{tabularx}
\usepackage{mdframed}

\newtheorem{theorem}{Theorem}
\newtheorem{lemma}{Lemma}
\newtheorem{assumption}{Assumption}

\begin{document}

\title{DFQ-ViT: Data-Free Quantization for Vision Transformers without Fine-tuning}


\author{Yujia Tong}
\email{tyjjjj@whut.edu.cn}

\author{Jingling Yuan}
\authornote{Corresponding author.}
\email{yjl@whut.edu.cn}

\author{Tian Zhang}
\email{zhangttt@whut.edu.cn}

\affiliation{%
  \institution{School of Computer Science and Artificial Intelligence, Wuhan University of Technology, Hubei Key Laboratory of Transportation Internet of Things}
  \city{Wuhan}
  \state{Hubei}
  \country{China}
}

\author{Jianquan Liu}
\email{jqliu@nec.com}
\affiliation{%
  \institution{NEC Corporation}
  \city{Tsukuba}
  \country{Japan}
}

\author{Chuang Hu}
\email{chuanghu@um.edu.mo}
\affiliation{%
  \institution{State Key Laboratory of Internet of Things for Smart City, University of Macau}
  \city{Macau}
  \country{China}
}

\renewcommand{\shortauthors}{Tong et al.}

\begin{abstract}
 Data-Free Quantization (DFQ) enables the quantization of Vision Transformers (ViTs) without requiring access to data, allowing for the deployment of ViTs on devices with limited resources. In DFQ, the quantization model must be calibrated using synthetic samples, making the quality of these synthetic samples crucial. Existing methods fail to fully capture and balance the global and local features within the samples, resulting in limited synthetic data quality. Moreover, we have found that during inference, there is a significant difference in the distributions of intermediate layer activations between the quantized and full-precision models. These issues lead to a severe performance degradation of the quantized model. To address these problems, we propose a pipeline for Data-Free Quantization for Vision Transformers (DFQ-ViT). Specifically, we synthesize samples in order of increasing difficulty, effectively enhancing the quality of synthetic data. During the calibration and inference stage, we introduce the activation correction matrix for the quantized model to align the intermediate layer activations with those of the full-precision model. Extensive experiments demonstrate that DFQ-ViT achieves remarkable superiority over existing DFQ methods and its performance is on par with models quantized through real data. For example, the performance of DeiT-T with 3-bit weights quantization is $4.29\%$ higher than the state-of-the-art. Our method eliminates the need for fine-tuning, which not only reduces computational overhead but also lowers the deployment barriers for edge devices. This characteristic aligns with the principles of Green Learning by improving energy efficiency and facilitating real-world applications in resource-constrained environments.
\end{abstract}




\maketitle

\section{Introduction}

Over the past few years, the area of Natural Language Processing (NLP) has experienced notable  advancements, where Transformer-based architectures like BERT~\cite{devlin2019bert} and GPT~\cite{brown2020language} have demonstrated exceptional capabilities in tasks such as language understanding and generation. This success has also propelled the application of Transformer architectures in the field of computer vision, giving rise to Vision Transformers (ViTs) \cite{dosovitskiy2020image}. ViTs have demonstrated impressive capabilities in various vision tasks, including image classification~\cite{he2016deep,tan2022fine,liu2021medical} and object detection~\cite{carion2020end,zhu2020deformable,xu2021exploring}, becoming a research hotspot in computer vision. However, the extensive parameter counts and high computational demands of these models pose significant challenges for practical deployment. Take the ViT-L model as an example, which has 307 million parameters and requires 64 billion floating-point operations (FLOPs) for inference~\cite{liu2021post}. Such substantial computational and memory requirements make it extremely difficult to deploy these models on hardware with limited resources, severely restricting their applications in edge computing and mobile devices.

To address these challenges, model compression techniques \cite{frantar2023sparsegpt,li2024selective,saha2023matrix} have become a focal point of research, with quantization~\cite{krishnamoorthi2018quantizing,gholami2022survey,li2016ternary} emerging as a particularly promising approach. Quantization reduces the bit-width of model weights and activations, thereby significantly decreasing memory usage and inference latency without substantially compromising model performance. However, the quantization process inevitably introduces quantization errors, which can lead to accuracy degradation. To mitigate this issue, traditional methods often rely on retraining the model with the original training data to adjust the quantized weights and restore accuracy. However, this approach faces two major obstacles in practical applications: First, access to the original training data is frequently limited due to strict privacy demands, such as those associated with medical imaging data. Second, retraining requires substantial computational resources and time, which is impractical for resource-constrained edge devices. Thus, the challenge of how to conduct efficient model quantization in the absence of original data has emerged as an urgent issue that needs to be addressed.

\begin{figure}[t]
\centering

\includegraphics[width=0.6\columnwidth]{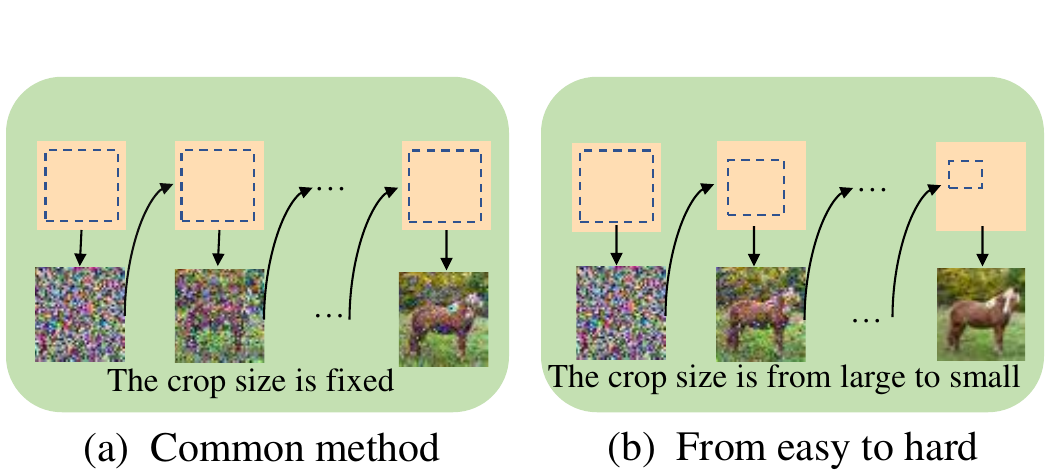} 
\caption{Comparison of sample synthesis strategies. (a) Existing methods use fixed maximum crop size which primarily captures global features but lacks local details. (b) Our proposed Easy to Hard (E2H) strategy progressively reduces crop size  during synthesis iterations. The dynamic cropping first establishes object contours through large crops (easy tasks) then refines local details via smaller crops (hard tasks). This curriculum learning approach better balances global-local feature learning compared to fixed-crop methods.
}
\label{fig1}
\end{figure}

To tackle these challenges, Data-Free Quantization (DFQ)~\cite{li2022patch} has emerged as a groundbreaking solution that eliminates the need for original training data. By generating synthetic samples to calibrate the quantized model, DFQ offers a promising alternative for scenarios where access to the original data is restricted. However, the effectiveness of DFQ hinges critically on the quality of the synthetic samples, which directly impacts the model's calibration and overall performance. Existing DFQ methods~\cite{cai2020zeroq,zhong2022intraq,qian2023rethinking} have primarily focused on Convolutional Neural Networks (CNNs), leveraging Batch Normalization (BN) to align synthetic samples with real data distributions. This approach is effective for CNNs but is fundamentally unsuitable for ViTs, which rely on Layer Normalization (LN) and do not store data statistics. To address this gap, PSAQ-ViT~\cite{li2022patch} was proposed, using patch similarity to extract random input data from a standard Gaussian distribution for iterative updates to adapt to the real data distribution. However, this method overlooks the importance of the order and difficulty in sample generation, as illustrated in Figure \ref{fig1}. This oversight can lead to synthetic samples that fail to fully capture the global features and local details of the data, thereby limiting the model's ability to generalize effectively. Moreover, even after calibrating the quantized model with synthetic samples, we observe that the intermediate layer activation distributions during inference still exhibit significant discrepancies compared to those of the full-precision model. These discrepancies accumulate across layers, ultimately resulting in a decline in the final performance of the quantized model.

In this paper, we introduce a novel Data-Free Quantization pipeline  for Vision Transformers (DFQ-ViT), designed to overcome the shortcomings of current approaches and markedly improve the performance of quantized models when original data is unavailable. Our approach introduces an \textbf{Easy to Hard (E2H)} strategy for sample generation, which begins with large crops to capture the global structures of objects and progressively refines the samples with smaller crops to enhance local details. This method not only improves the quality of synthetic samples but also better simulates the complexity of real data, leading to more effective model calibration. Additionally, we introduce an \textbf{Activation Correction Matrix (ACM)} to adjust the intermediate layer activations during inference, aligning them more closely with the performance of the full-precision model. By correcting the intermediate activations, we can reduce the effect of quantization errors that build up across layers, which in turn improves the overall performance and reliability of the quantized model. Through extensive experiments, we demonstrate that DFQ-ViT outperforms existing data-free quantization methods and achieves performance comparable to models quantized with real data. Notably, our framework completely eliminates the fine-tuning process, thus avoiding the substantial computational cost associated with retraining. This makes the quantization process more energy-efficient and environmentally friendly. Moreover, it significantly reduces the technical barriers to deploying ViTs on edge devices. These characteristics make DFQ-ViT fully aligned with the core goals of green learning—achieving high computational efficiency while reducing  deployment costs in practical applications. Our main contributions are as follows:
\begin{itemize}
 \item We introduce the Easy to Hard (E2H) strategy into the sample synthesis process, generating samples in an easy-to-hard order, which improves the quality of synthetic samples. Moreover, we theoretically prove the effectiveness of the E2H strategy.
 \item We introduce the Activation Correction Matrix (ACM) to correct the intermediate layer activations of the quantized model during inference, reducing the intermediate layer quantization errors and thereby enhancing the performance of the quantized model.
 \item We conduct extensive experiments to verify that DFQ-ViT is significantly better than existing DFQ methods, and its performance is comparable to models quantized with real data.
 \end{itemize}

\section{Related Works}
In this section, we initially review the current methods for  \textit{Vision Transformer compression}, and subsequently delve into the primary research focus of this paper: \textit{data-free quantization.}

\subsection{Vision Transformer Compression.}
Vision Transformers (ViTs) have achieved remarkable results in computer vision. Nevertheless, the extensive parameters and the significant computational demands of ViTs during inference pose challenges for deployment on resource-constrained hardware. Therefore, compressing ViTs has become an important research area. Existing compression methods mainly include pruning, distillation, low-rank decomposition, and quantization. Pruning \cite{Yang_2023_CVPR, yu2023unified} reduces the size of ViTs by removing unimportant components. Distillation \cite{wu2022tinyvit,yang2024vitkd} transfers the knowledge from a larger teacher model to a smaller student model. Low-rank decomposition \cite{dong2024low} saves storage space by decomposing a large matrix into smaller matrices. Quantization \cite{hooper2024kvquant,lee2024owq,dettmers2022gpt3} is an efficient model compression technique that reduces the storage requirements and computational complexity of models by compressing weights and activations from floating-point numbers to low-bit integers, thereby accelerating inference speed. Depending on whether retraining or fine-tuning is involved, quantization methods are divided into two main categories: Quantization-Aware Training (QAT)~\cite{zhou2016dorefa,esser2019learned,li2022q} and Post-Training Quantization (PTQ)~\cite{liu2021post,lin2021fq,yuan2021ptq4vit}.
QAT incorporates quantization operations during training. By retraining the quantized model on the original dataset, it mitigates performance degradation caused by low-bit quantization. QAT can simulate quantization errors and adjust model parameters to maintain high accuracy but requires substantial computational resources and time, making it impractical for resource-constrained devices.
PTQ directly quantizes pre-trained full-precision models without retraining. It only needs a small amount of data for calibration, enabling rapid deployment. PTQ is more suitable for resource-constrained devices as it avoids the high costs of retraining. However, its performance improvement is usually less significant than QAT's since it cannot optimize parameters through training. Nonetheless, PTQ remains attractive for scenarios requiring quick deployment and resource optimization.

Despite the respective advantages of QAT and PTQ in different application scenarios, both face a common challenge: dependence on original data. Whether through retraining or calibration, these methods require access to the original training data. However, in many practical applications, original data may not be available due to privacy protection, data security, or data ownership issues. For example, in fields such as medical imaging and financial data, the sensitivity of the data makes it impossible to use for model retraining or calibration. Therefore, developing a quantization method that does not rely on original data is of great significance for solving model deployment problems in these scenarios.

\subsection{Data-Free Quantization.}
In response to the issue of inaccessible data, Data-Free Quantization (DFQ)\cite{banner2019post,cai2020zeroq,xu2020generative} has been proposed. The fundamental concept of DFQ involves creating synthetic samples as a substitute for the original data to calibrate the quantized model. This method completely eliminates the dependence on original data, allowing the quantization process to be carried out in scenarios where data is restricted or unavailable. In recent years, DFQ\cite{zhang2021diversifying} has made significant progress in the field of deep neural networks and has attracted the attention of many researchers.
PSAQ-ViT~\cite{li2022patch} is  the first to apply data-free quantization to Vision Transformers (ViTs), which extracts random input data from a standard Gaussian distribution using patch similarity and iteratively updates to adapt to the real data distribution, thereby generating synthetic samples. This method first introduced data-free quantization to ViTs, providing a new approach to solving the quantization problem of ViTs. However, PSAQ-ViT ignores the order and difficulty of sample generation, which may lead to the generated samples failing to fully capture the global features and local details of the data, thereby limiting the calibration effect of the model.
To further improve the performance of data-free quantization, PSAQ-ViT V2~\cite{li2023psaq} introduced an adaptive teacher-student strategy. Under the supervision of the full-precision model, the generated samples and the quantized model continuously evolve through competition and interaction in a cyclical manner. This method significantly improves the accuracy of the quantized model, but its drawback is the need to retrain the quantized model. Retraining not only requires significant computational resources and time but may also introduce additional errors, making PSAQ-ViT V2 difficult to widely apply in resource-constrained scenarios.

Quantizing ViTs in resource-constrained scenarios where data is unavailable is a highly challenging problem. Existing data-free quantization methods have to some extent solved the problem of inaccessible data, but most of them rely on retraining or fine-tuning, making them difficult to implement on resource-constrained devices. Moreover, these methods often ignore the global features and local details of samples when generating synthetic samples, resulting in limited calibration effects. To address these issues, we propose a data-free quantization pipeline designed for ViTs—DFQ-ViT. DFQ-ViT introduces an Easy to Hard (E2H) sample generation strategy and an Activation Correction Matrix (ACM) to improve the accuracy of the quantized model without retraining or fine-tuning. Our method not only enhances the quality of synthetic samples but also reduces the impact of quantization errors on the final results by correcting the activations of intermediate layers, thereby achieving efficient and reliable model quantization in resource-constrained scenarios where data is unavailable.

\begin{figure*}[t]
\centering
\includegraphics[width=1\textwidth]{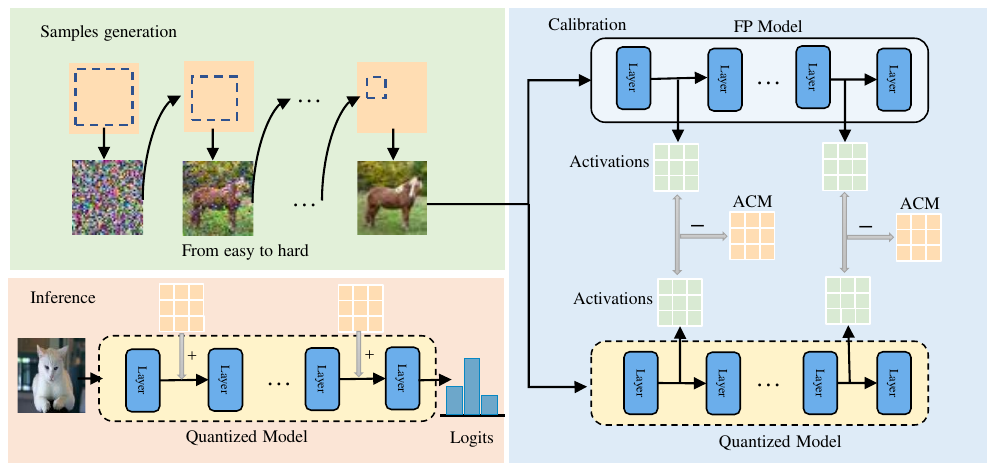} 
\caption{The overview of DFQ-ViT. During the sample generation phase, we consider the difficulty and order of synthetic samples, generating samples from easy to difficult, enabling the samples to capture and balance the global and local features in the samples. During the model calibration phase, we calculate and store the intermediate layer's activation correction matrix to align the intermediate layer activations of the quantized model and the full-precision model during inference. }
\label{fig2}
\end{figure*}

\section{Methodology}
In this section, we first introduce the computational mechanisms of ViTs and provide an overview of the fundamental principles of data-free quantization (DFQ). Then, we detail the Easy to Hard sample synthesis strategy. In addition, to reduce the quantization errors of the model, we introduce the Activation Correction Matrix. Figure \ref{fig2} provides an overview of DFQ-ViT.

\subsection{Preliminaries}
In the context of a typical transformer, the input is formatted as a series of vectors. The workflow commences with the segmentation of the input image into a number of patches with a consistent size.  Each patch is linearly projected to vector. These vectors, representing the tokens, are fed into the vision transformer, denoted as \( X \in \mathbb{R}^{N \times D} \), where \( N \) represents the number of tokens and \( D \) indicates the hidden size. The vectors \( X \) must go through multiple transformer blocks. Each block is composed of a multi-head self-attention (MSA) module and a multi-layer perceptron (MLP) module. In each head, the attention weight is computed based on the  \( Q_i = X W^{Q}_i \),  \( K_i = X W^{K}_i \), and  \( V_i = X W^{V}_i \), and is given by:
\begin{equation}
  \text{Att}_i(Q_i, K_i, V_i) = \text{softmax}\left(\frac{Q_i K^T_i}{\sqrt{d}}\right) V_i,
\label{eq1}
\end{equation}
where \( d \) denotes the hidden size of each head. The outputs from each head are aggregated by concatenation to create the output of the MSA:
\begin{equation}
  \text{MSA}(X) = \text{Concat}(\text{Att}_1, \text{Att}_2, \ldots, \text{Att}_i),
\label{eq2}
\end{equation}
where \( i \) represents the number of heads. Subsequently, the output is passed into the MLP, which is composed of two fully connected layers.

The significant computational resource consumption in the vision transformer is primarily due to the matrix multiplication operations in MSA and MLP. Quantizing the parameters involved in these matrix multiplications can effectively reduce the computational overhead and storage requirements of the vision transformer. We utilize the commonly employed uniform symmetric quantization technique. Given a scaling factor \( \Delta \), a floating-point value \( x \) is quantized to a \( k \)-bit integer value \( x_q \) according to:
\begin{equation}
    x_q = \text{clip}\left(\text{round}\left(\frac{x}{\Delta}\right), -2^{k-1}, 2^{k-1} - 1\right),
\label{eq3}
\end{equation}
where \( \text{round}(\cdot) \) denotes rounding to nearest integer, and \( \text{clip}(\cdot) \)  restricts the output within the range representable by \( k \)-bit integers.

DFQ necessitates synthetic samples to calibrate the quantized model. Most prior works leverage the statistical information in the BN layer of the full-precision model for sample synthesis. However, since ViTs do not have BN layers, these previous methods are not suitable for DFQ of ViTs. To address this issue, PSAQ-ViT introduces Patch Similarity Metric for sample synthesis. For the \( l \)-th layer of the ViTs, the cosine similarity between the outputs of the MSA module is computed as:
\begin{equation}
\Gamma_l(u_i, u_j) = \frac{u_i \cdot u_j}{\|u_i\| \|u_j\|},
\label{eq4}
\end{equation}
where \( u_i \) and \( u_j \) are vectors in the patch dimension. 
The diversity of patch similarity is assessed using differential entropy that is computed through kernel density estimation, given by:
\begin{equation}
D_{ps}^{l} = -\int \hat{f}(x) \log \hat{f}(x) \, dx,
\label{eq5}
\end{equation}
where \( \hat{f}(x) \)  represents the continuous probability density function derived via kernel density estimation:
\begin{equation}
\hat{f}(x) = \frac{1}{Mh} \sum_{m=1}^M K\left(\frac{x - x_m}{h}\right),
\label{eq6}
\end{equation}
where \( K(\cdot) \) denotes the kernel function, \( h \) represents the bandwidth, \( x_m \) ( \( m \in \{1, \ldots, M\} \) ) is a sample point selected from \( \Gamma_l \) and acts as the kernel center, and \( x \) refers to the specified test point. Patch Similarity Entropy Loss is  the aggregate sum of  the differential entropies across all layers, quantifying the overall diversity of patch similarity:
\begin{equation}
L_{PSE} = -\sum_{l=1}^L D_{ps}^{l},
\label{eq7}
\end{equation}
The One-hot loss \( L_{OH} = CE(P(I), c) \) encourages synthetic images to be predicted within a predefined category \( c \). Here, \( CE \) denotes the Cross-Entropy Loss, which  quantifies the difference between the model's predictions \( P(I) \) for an image \( I \) and the true class labels \( c \). The Total Variance Loss \( L_{TV} = \int \int |\nabla I(\tau_1, \tau_2)| \, d\tau_1 \, d\tau_2 \) acts as a regularization term to ensure pixel-level smoothing of the generated images. It involves integrating the absolute value of the gradient \( \nabla I(\tau_1, \tau_2) \) of the image \( I \) over the entire image domain. Therefore, the final loss for sample generation is:
\begin{equation}
L_{TOTAL} = L_{PSE} + \alpha L_{OH} + \beta L_{TV}.
\label{eq8}
\end{equation}

\subsection{Easy to Hard}

DFQ requires the generation of synthetic samples to calibrate the quantized model, and thus the quality of these synthetic samples directly determines the performance of the quantized model. High-quality synthetic samples can better approximate the distribution of real data, thereby helping the quantized model maintain higher accuracy in low-precision environments. However, existing methods (such as PSAQ-ViT) have an obvious limitation: they fail to consider the difficulty level of sample synthesis. Global features (e.g., the overall contours and spatial layout of objects) are relatively easy to synthesize (defined as \textbf{"easy"} tasks) due to their stronger semantic saliency in visual tasks. In contrast, local features (e.g., texture details and edge information) require precise pixel-level alignment and are significantly more difficult to synthesize (defined as \textbf{"hard"} tasks). This difference in difficulty stems from a fundamental principle of human visual perception — we always first recognize the overall shape of an object before gradually observing its local details. The oversight of existing methods makes it difficult for the model to balance global and local features during the synthesis process, thereby affecting the overall quality of the synthetic samples.

To address this issue, we draw inspiration from Curriculum Learning \cite{wang2021survey,graves2017automated,hacohen2019power} and propose an \textbf{Easy to Hard (E2H)} sample synthesis strategy. Curriculum Learning is a training approach that emulates the way humans learn, with the core idea of starting the model with simpler tasks or samples and gradually transitioning to more complex ones. This strategy has been proven to significantly improve the learning efficiency and generalization ability of models~\cite{yin2023dataset}. The E2H strategy is based on this idea, guiding the synthesis of samples from global to local features by progressively adjusting the cropping ratio during the synthesis process, thereby enhancing the quality of synthetic samples.

In the early stages of synthesis, the E2H strategy employs a larger cropping area to outline the rough contours of objects, thereby establishing a global structural foundation for the samples. The goal of this stage is to generate the overall features of the samples, such as the general shape, position, and layout of objects. As the synthesis process progresses, the E2H strategy gradually reduces the size of the cropping area, focusing on refining and enhancing local details of the image, such as textures, edges, and local features. This progressive shift from global to local features enables the samples to more effectively balance the acquisition of global and local features, which in turn notably enhances the quality of the synthetic samples.

In practical implementation, we can utilize the {\em RandomResizedCrop} function to perform cropping on the sample's image before each iteration. As the iterations progress, we gradually decrease the size of the crop. The specific process is shown in Algorithm \ref{alg1}. The initial sample \( x_0 \) is drawn from a Gaussian distribution \( \mathcal{N}(\mu, \sigma^2) \). At \( t = 0 \), the cropping ratio \( \delta_0 \) is close to \( \delta_u \). According to the formula:
\begin{equation}
\delta_0 = \delta_l + (\delta_u - \delta_l) \cdot \left(1 + \cos\left(\pi \cdot \frac{0}{T}\right)\right) / 2 = \delta_u.
\label{eq9}
\end{equation}
Therefore, \( \delta_0 \gets \delta_u \), which allows the model to interact with a larger range of image information, helping it learn global features. When \( t = T \), the cropping ratio \( \delta_T \) approaches \( \delta_l \). According to the formula:
\begin{equation}
\delta_T = \delta_l + (\delta_u - \delta_l) \cdot \left(1 + \cos\left(\pi \cdot \frac{T}{T}\right)\right) / 2 = \delta_l.
\label{eq10}
\end{equation}
Therefore, \( \delta_T \gets \delta_l \), meaning the model focuses more on smaller ranges within the image details.

\begin{algorithm}[tb]
\large
\caption{Generate Synthetic Sample from Easy to Hard via \emph{RandomResizedCrop}}
\label{alg1}
\begin{algorithmic}[1] 
\REQUIRE full-precision vision transformer model $\theta$, total generation iterations $T$, lower and upper bounds of crop scale $\delta_{l}$ and $\delta_\text{u}$.\\
\ENSURE Synthetic sample $x$.
\STATE Initialize $x_0$ from a Gaussian distribution.
\FOR{$t \in [0, \ldots, T-1]$} 
    \STATE $\delta_t = \delta_l + (\delta_u - \delta_l) * \left(1 + \cos\left(\pi * t / T\right)\right) / 2$ 
    \STATE $\hat{x_t} \gets$ \emph{RandomResizedCrop}$(x_t, \texttt{min\_crop}=\delta_t, \texttt{max\_crop}=\delta_u)$ 
    \STATE Optimize $\hat{x_t}$ to obtain $\hat{x'_t}$ using Eq.(\ref{eq8})
    \STATE Rescale $\hat{x'_t}$ to the size of $x_t$ to obtain $x_{t+1}$
\ENDFOR 
\RETURN $x_{T}$
\end{algorithmic}
\end{algorithm}

In this way, the E2H  strategy guides the initial sample to learn from global to local features step by step during the iteration process, thereby improving the quality of the synthesized samples. Taking the synthesis of images in the "kite" category as an example, when the cropping size is large, the iterations at this stage are relatively simple because the sample only needs to synthesize the preliminary outline and general layout of the kite. In contrast, when the cropping size is small, the task becomes more complex, as the sample needs to synthesize more details of the kite, such as its tail and feathers. The core idea of the E2H strategy, which gradually transitions from simpler tasks (larger cropping) to more complex tasks (smaller cropping), has already been proven to be effective in prior work on dataset distillation ~\cite{yin2023dataset}. We introduce this idea into the process of synthetic sample generation for data-free quantization and theoretically analyze its effectiveness.

\subsection{Theoretical Analysis of Easy to Hard Strategy}
We theoretically analyze the impact of the E2H strategy and the conventional strategy (Fixed strategy) on the quality of synthetic samples.

\subsubsection{Problem Formulation}
Consider the example synthesis problem with dynamic complexity control. We define the composite loss function:
\begin{equation}
L_{\text{Total}}(x; \delta) = L_{\text{PSE}}(x) + \alpha L_{\text{OH}}(x) + \beta L_{\text{TV}}(x)
\end{equation}
where $\delta \in [\delta_{\min}, \delta_{\max}]$ controls the data complexity. We compare two strategies:
\begin{itemize}
\item \textbf{Fixed Strategy}: $\delta_t = \delta_{\max},\ \forall t$
\item \textbf{E2H Strategy}: Monotonically decreasing sequence $\{\delta_t\}_{t=0}^T$ with $\delta_{t+1} < \delta_t$
\end{itemize}

\subsubsection{Key Assumptions} We make the following assumptions about the loss function $L_{\text{Total}}(x; \delta)$:
\begin{assumption}[Lipschitz Smoothness]
For any $x$ and $\delta_t, \delta_{t+1}$, there exists a constant K:
\begin{equation}
\|\nabla_x L(x; \delta_t) - \nabla_x L(x; \delta_{t+1})\| \leq K |\delta_t - \delta_{t+1}|
\end{equation}
\end{assumption}

\begin{assumption}[$\beta$-Smoothness]
For any $x,x'$ and $\delta$, there exists a constant $\beta$ :
\begin{equation}
L(x'; \delta) \leq L(x; \delta) + \nabla_x L(x; \delta)^\top (x' - x) + \frac{\beta}{2} \|x' - x\|^2
\end{equation}
\end{assumption}

\begin{assumption}[Bounded Gradient]
There exists $G > 0$ such that for all $x$ and $\delta$:
\begin{equation}
\|\nabla_x L(x; \delta)\| \leq G
\end{equation}
\end{assumption}

\begin{assumption}[Cumulative Gradient]
For any $T > 0$, the cumulative gradient of E2H satisfies:
\begin{equation}
\sum_{t=0}^{T-1} \|\nabla_x L(x_t^{E2H}; \delta_t)\|^2 \geq \sum_{t=0}^{T-1} \|\nabla_x L(x_t^{Fixed}; \delta_{\max})\|^2 + \frac{KG\Delta}{\eta - \frac{\beta\eta^2}{2}}
\end{equation}
where $\Delta = \delta_{\max} - \delta_{\min}$, $\eta$ is learning rate.
\end{assumption}

\subsubsection{Main Theoretical Results}
\begin{lemma}[Single-Step Optimization Bound for Fixed] For the Fixed strategy with learning rate $\eta$:
\begin{equation}
L(x_{t+1}; \delta_{\max}) \leq L(x_t; \delta_{\max}) - \left(\eta - \frac{\beta\eta^2}{2}\right) \|\nabla_x L(x_t; \delta_{\max})\|^2.
\end{equation}
\end{lemma}

\begin{proof}
The fixed strategy always uses $\delta_{\max}$, and its update rule is:
\begin{equation}
x_{t+1} = x_t - \eta \nabla_x L(x_t; \delta_{\max}).
\end{equation}

Since $\delta_t = \delta_{t+1} = \delta_{\max}$, according to Assumption 2, the single-step loss satisfies:
\begin{equation}
L(x_{t+1}; \delta_{\max}) \leq L(x_t; \delta_{\max}) + \nabla_x L(x; \delta_{\max})^\top (x_{t+1} - x_{t})  + \frac{\beta}{2} \|x_{t+1} - x_{t}\|^2 .
\end{equation}

Combining the above two equations yields Lemma 1.
\end{proof}

\begin{lemma}[Single-Step Optimization Bound for E2H]
For the E2H strategy with learning rate $\eta$:
\begin{equation}
L(x_{t+1}; \delta_{t+1}) \leq L(x_t; \delta_t) - \left(\eta - \frac{\beta\eta^2}{2}\right)\|\nabla_x L(x_t; \delta_t)\|^2 + K|\delta_t - \delta_{t+1}|\cdot \|x_{t+1} - x_t\|
\end{equation}
\end{lemma}

\begin{proof}
When $\delta = \delta_t$ is fixed, we have:
\begin{equation}
L(x_{t+1}; \delta_t) \leq L(x_t; \delta_t) - \left(\eta - \frac{\beta\eta^2}{2}\right) \|\nabla_x L(x_t; \delta_t)\|^2.
\end{equation}

Since $\delta$ changes from $\delta_t$ to $\delta_{t+1}$, according to Assumption 1, combining the mean value theorem, consider the impact of the change in $\delta$ on the loss function:
\begin{equation}
L(x_{t+1}; \delta_{t+1}) \leq L(x_{t+1}; \delta_t) + K |\delta_t - \delta_{t+1}| \cdot \|x_{t+1} - x_t\|.
\end{equation}

Combining the above two equations yields Lemma 2.
\end{proof}

\begin{theorem}[Superiority of E2H Strategy]
Under Assumptions 1-4, let both strategies start from the same initial point $x_0$, and let the learning rate satisfy $\eta \leq \frac{2}{\beta}$. The E2H strategy achieves better final performance:
\begin{equation}
L(x_T^{E2H}; \delta_T) \leq L(x_T^{Fixed}; \delta_T)
\end{equation}
\end{theorem}

\begin{proof}
For the Fixed strategy, summing Lemma 1 over $T$ steps:
\begin{equation}
L(x_T^{Fixed}; \delta_T) \leq L_0 - \left(\eta - \frac{\beta\eta^2}{2}\right)\sum_{t=0}^{T-1} \|\nabla_x L(x_t^{Fixed}; \delta_{\max})\|^2
\end{equation}
where $ L_0 $ represents the loss value of both strategies (\text{E2H} and \text{Fixed}) at the initial time step $ t = 0 $.

For the E2H strategy, summing Lemma 2 over $T$ steps and using $\|x_{t+1}-x_t\| \leq \eta G$ (from Assumption 3):
\begin{align}
L(x_T^{E2H}; \delta_T) &\leq L_0 - \left(\eta - \frac{\beta\eta^2}{2}\right)\sum_{t=0}^{T-1} \|\nabla_x L(x_t^{E2H}; \delta_t)\|^2 + K\eta G\sum_{t=0}^{T-1}|\delta_t - \delta_{t+1}| \\
&\leq L_0 - \left(\eta - \frac{\beta\eta^2}{2}\right)\sum_{t=0}^{T-1} \|\nabla_x L(x_t^{E2H}; \delta_t)\|^2 + K\eta G\Delta
\end{align}

Substituting the gradient advantage condition:
\begin{align}
L(x_T^{E2H}; \delta_T) &\leq L_0 - \left(\eta - \frac{\beta\eta^2}{2}\right)\left[\sum_{t=0}^{T-1} \|\nabla_x L(x_t^{Fixed}; \delta_{\max})\|^2 + \frac{KG\Delta}{\eta - \frac{\beta\eta^2}{2}}\right] + K\eta G\Delta \\
&= \left[L_0 - \left(\eta - \frac{\beta\eta^2}{2}\right)\sum_{t=0}^{T-1} \|\nabla_x L(x_t^{Fixed}; \delta_{\max})\|^2\right] - KG\Delta + K\eta G\Delta \\
&\leq L(x_T^{Fixed}; \delta_T) - KG\Delta(1 - \eta)
\end{align}
When $\eta < 1$, the final term $-KG\Delta(1-\eta)$ is negative, proving:
\begin{equation}
L(x_T^{E2H}; \delta_T) \leq L(x_T^{Fixed}; \delta_T)
\end{equation}

\end{proof}

This theoretical result explains why E2H can generate higher-quality samples than fixed-complexity approaches, particularly in high-dimensional vision transformer quantization scenarios where both global structure and local details need precise recovery.

\subsection{Activation Correction Matrix}
Quantization is one of the key techniques for deploying deep learning models to hardware platforms. By transforming the  weights and activations from floating-point representations to low-bit-width quantized representations, quantization can substantially decrease the model's computational complexity and memory requirements. This characteristic  renders quantized models more appropriate for deployment in resource-constrained hardware environments, like mobile devices and edge computing devices. However, the quantization process inevitably introduces quantization errors, which may negatively impact the model's accuracy and performance. In some cases, the performance of the quantized model may drop significantly, thereby limiting its usability in practical applications.

To address this issue, previous research has attempted to reduce performance loss by retraining the quantized models. Specifically, these methods typically use synthetic samples to fine-tune the quantized models in order to recover the precision lost due to quantization. Although this approach can be effective in some cases, retraining models in environments with limited hardware resources face many challenges. For example, retraining may require substantial computational resources and time, which are difficult to achieve on resource-constrained devices.

To address these challenges, this paper proposes a novel forward calibration method. Unlike traditional retraining approaches, our method is a one-time calibration process that does not require retraining the model. The core idea of this method is to use a small number of synthetic samples to calibrate the quantized model by adjusting the intermediate layer activations through the forward propagation process, making them closer to the activation patterns of the full-precision model. This approach not only saves a significant amount of computational resources but also quickly adapts to the needs of hardware deployment, making it particularly suitable for use in resource-constrained environments.

Specifically, the key to improving model performance lies in reducing quantization errors, which primarily stems from the discretized representation of model weights and activation functions. During the quantization process, the continuous floating-point values are mapped to a finite set of discrete values, inevitably introducing errors. These errors can accumulate as the number of layers increases in deep networks, ultimately having a significant impact on the model's final performance. For example, in some deep convolutional neural networks (CNNs), the accumulation of quantization errors can lead to a substantial drop in classification accuracy. Inspired by the bias correction technique introduced by~\citeauthor{nagel2019data}~\citeyearpar{nagel2019data} in the DFQ  method for CNNs, we found that correcting the intermediate layer activations in Vision Transformers (ViTs) can also effectively reduce quantization errors. Based on this finding, we propose a forward calibration method suitable for ViTs. This method introduces an Activation Correction Matrix (ACM) to adjust the intermediate layer activations of the quantized model, bringing them closer to the activation patterns of the full-precision model. This approach significantly improves the performance of the quantized model without adding extra computational burden, making it more accurate and reliable for hardware deployment.

Essentially, a classification network can be simply viewed as a mapping from input \( x \) to output \( p(w, x) \) through the function \( f(w, x) \). We denote the intermediate layer activations of the full-precision model as \( p_{\mathbf{T}}^{i}(w^r,x) \), and those of the quantized model as \( p_{\mathbf{Q}}^{i}(w^q,x) \). To avoid the accumulation of errors introduced by quantization as the number of layers increases, an intuitive idea is to align the intermediate layer activations of the quantized model with those of the full-precision model, satisfying the following formula:
\begin{equation}
    \min \mathbf{distance}(p_{\mathbf{Q}}^{i}(w^q,x), p_{\mathbf{T}}^{i}(w^r,x)),
\label{eq11}
\end{equation}
To achieve the aforementioned alignment, we introduce an activation correction matrix for the intermediate layer activations of the quantized model, represented as:
\begin{equation}
    \mathbf{Matrix}_{ACM}^{i} = \frac{1}{N} \sum_{n=1}^N [p_{\mathbf{T}}^{i}(w^r,x) - p_{\mathbf{Q}}^{i}(w^q,x)],
\label{eq12}
\end{equation}
where \( N \) represents the number of synthetic samples used for calibration. During the inference phase, the forward process of the quantized model is:
\begin{equation}
    x^{i-\gamma} = p_{\mathbf{Q}}^{i-\gamma}(w^q,x) + \mathbf{Matrix}_{ACM}^{i-\gamma},
\label{eq13}
\end{equation}
\begin{equation}
    x^{i} = p_{\mathbf{Q}}^{i}(w^q,x^{i-\gamma}) + \mathbf{Matrix}_{ACM}^{i}.
\label{eq14}
\end{equation}
where \( x \) denotes the original input to the model, \( x^{i} \) indicates the input to the \( i \)-th layer of the quantized model during the inference phase, and \( \gamma \) represents the interval at which an activation correction matrix is introduced every \( \gamma \) layer. By introducing the activation correction matrix, our method can effectively reduce quantization errors and improve the performance of the quantized model without retraining, making it particularly suitable for environments with limited hardware resources.

\begin{algorithm}[t]
\large
\caption{The overall pipeline of DFQ-ViT}
\label{alg2}
\begin{algorithmic}[1] 
\REQUIRE full-precision vision transformer model $\theta$, generation iterations $T$, lower and upper bounds of crop scale $\delta_{l}$ and $\delta_\text{u}$ in \emph{RandomResizedCrop}.\\
\ENSURE Quantized weights $\hat{\theta}$. 
\STATE Initialize $x_0$ from a Gaussian distribution; 
\STATE Synthesize calibration samples $x$ following the procedure of Algorithm \ref{alg1};
\STATE Calibrate quantized model $\hat{\theta}$ by $x$;
\STATE Obtain activation correction matrix by  Eq.(\ref{eq12}) ;
\STATE Calculate output when inference following Eq.(\ref{eq13}) and Eq.(\ref{eq14}).
\end{algorithmic}
\end{algorithm}

DFQ-ViT primarily comprises three phases: the sample generation phase, the model calibration phase, and the inference phase. The overall pipeline is shown in Algorithm \ref{alg2}. In the sample generation phase, we consider the difficulty and order of synthetic samples, generating samples from easy to difficult and enabling the samples to capture and balance the global and local features in the samples, thereby improving sample quality. In the model calibration phase, we calculate and store the intermediate layer's activation correction matrix to align the intermediate layer activations of the quantized model and the full-precision model during the inference phase, effectively avoiding the accumulation of errors introduced by quantization as the number of layers increases. ViTs segment the image into patches and model global relationships through self-attention mechanisms, with its performance highly dependent on global context. E2H directly adapts to ViTs's patch processing mode by progressively generating from large crops (global contours) to small crops (local details). In contrast, CNNs extract hierarchical features through the local receptive fields of convolution kernels, without the need to explicitly distinguish the generation order of global and local features. Additionally, ViTs involve more matrix multiplications and are more sensitive to quantization errors, requiring targeted correction by ACM. Therefore, E2H and ACM are more effective for ViTs, and DFQ-ViT is a data-free quantization pipeline specifically designed for ViTs.

\section{Experiments}
\subsection{Experimental Settings}
\textbf{Models and Datasets.} To validate DFQ-ViT, we select various popular vision transformer models, including DeiT-T/S/B~\cite{touvron2021training} and Swin-T/S~\cite{liu2021swin}. Given that ViT and DeiT share the same architecture, we omit the evaluation of ViT. For the dataset, we chose ImageNet-1k~\cite{deng2009imagenet}, which contains approximately 1.2 million training images and 50,000 validation images across 1,000 categories. This dataset is widely used for benchmarking vision models and provides a robust testbed for evaluating the quantization performance of DFQ-ViT. Additionally, we include the CIFAR-10 and CIFAR-100 datasets\cite{krizhevsky2009learning}. CIFAR-10 contains 60,000 color images divided into 10 classes, with 50,000 training images and 10,000 test images. CIFAR-100 includes 60,000 color images in 100 classes, split into 50,000 training images and 10,000 test images. Both datasets are frequently utilized for evaluating image classification tasks.

\noindent\textbf{Baselines.} Given the lack of prior work on Data-Free Quantization for Vision Transformers without fine-tuning, we use PSAQ-ViT~\cite{li2022patch} as our baseline. To evaluate our method comprehensively, we also establish baselines using real images and Gaussian noise for calibration. Please note that PSAQ-ViT V2 is not used as a baseline due to its requirement for retraining, which makes the quantization process difficult to carry out on resource-constrained devices.

\noindent\textbf{Evaluation Metrics.} We utilize Top-1 and Top-5 accuracy as our evaluation metrics. For inference, we report the Top-1 and Top-5 accuracy of the full-precision models employed. Table \ref{tab0} provides the Top-1 and Top-5 accuracy of various full-precision ViT models on the ImageNet-1K, CIFAR-10 and CIFAR-100 dataset.

\begin{table}[htbp]
\centering
\caption{The Top-1 and Top-5 accuracy of various full-precision ViT models on ImageNet, CIFAR-10, and CIFAR-100 datasets.}
\begin{tabular}{cccccccc}
\hline
\multirow{2}{*}{Model} & \multirow{2}{*}{bit-width} & \multicolumn{2}{c}{ImageNet-1K} & \multicolumn{2}{c}{CIFAR-10} & \multicolumn{2}{c}{CIFAR-100} \\
\cline{3-8}
& & Top-1 & Top-5 & Top-1 & Top-5 & Top-1 & Top-5 \\ \hline
DeiT-T  & \multirow{5}{*}{W32/A32} & 72.14 & 91.13 & 95.97 & 99.91 & 80.64 & 96.11 \\
DeiT-S  & & 79.83 & 94.95 & 97.55 & 99.94 & 86.28 & 97.88 \\
DeiT-B  & & 81.79 & 95.59 & 98.20 & 99.98 & 89.54 & 98.91 \\
Swin-T  & & 81.37 & 95.54 & 97.56 & 99.98 & 87.10 & 98.53 \\
Swin-S  & & 83.21 & 96.32 & 98.14 & 99.97 & 88.61 & 98.74 \\ \hline
\end{tabular}
\label{tab0}
\end{table}

\noindent\textbf{Implementation Details.} 
Following PSAQ-ViT~\cite{li2022patch}, we apply fundamental quantization parameter calibration. We use symmetric uniform quantization for weights and asymmetric uniform quantization for activations. Calibration involves 16 images, each processed for 500 iterations. During the image synthesis stage, for the DeiT series models, we utilize the Adam optimizer with a learning rate of 0.25. For the Swin Transformer series models, we employ the Adam optimizer with a learning rate of 0.2. Parameters are set as $\alpha = 1.0$, $\beta = 0.05$, $\delta_l = 0.08$, and $\delta_u = 1.0$.

\subsection{Performance Comparison}

Table \ref{tab1} presents the Top-1 Accuracy and Top-5 Accuracy of various models at different quantization bit-precisions on the ImageNet-1k dataset. In the data-free quantization scenario, it is possible to calibrate the quantized model using randomly generated Gaussian noise data. However, this method often results in poor performance of the quantized model due to its lack of capturing the structure and features of the data. For example, under the 3-bit weight quantization (W3/A8) setting for the DeiT-T model, the Top-1 accuracy of the model calibrated with Gaussian noise is only 3.90\%, and the Top-5 accuracy is 20.61\%. This indicates that calibrating the quantized model using Gaussian noise data is not feasible.

In addition to the comparison with the Gaussian noise calibration method, we also conducted a detailed comparison between DFQ-ViT and the existing data-free quantization method, PSAQ-ViT. The experimental results show that DFQ-ViT achieves significant performance improvements on various Vision Transformer models. For example, under the 3-bit weight quantization (W3/A8) setting for the DeiT-T model, DFQ-ViT achieves a Top-1 accuracy of 27.97\% and a Top-5 accuracy of 51.53\%, which are 4.29\% and 5.30\% higher than those of PSAQ-ViT, respectively. This improvement indicates that DFQ-ViT can better preserve the original performance of the model in low-bit quantization scenarios. Moreover, under higher quantization precisions (such as W4/A8, W6/A8, and W8/A8), DFQ-ViT continues to outperform PSAQ-ViT, further demonstrating its robustness and effectiveness across different quantization levels. These results show that by introducing the “Easy to Hard” (E2H) sample synthesis strategy and the Activation Correction Matrix (ACM), DFQ-ViT can significantly enhance the quality of synthetic samples and effectively align the intermediate layer activations between the quantized model and the full-precision model, thereby achieving efficient model quantization without the need for real data. 

To further validate the performance of DFQ-ViT, we also compared it with quantization methods that use real data for calibration. The experimental results show that, under certain quantization precisions, the performance of DFQ-ViT even surpasses that of models calibrated with real data. For example, under the 4-bit weight quantization (W4/A8) setting for the DeiT-T model, DFQ-ViT achieves a Top-1 accuracy of 65.84\% and a Top-5 accuracy of 87.04\%, while the model calibrated with real data has a Top-1 accuracy of 65.38\% and a Top-5 accuracy of 86.74\%. DFQ-ViT outperforms the real-data-calibrated model by 0.46\% in Top-1 accuracy and by 0.30\% in Top-5 accuracy. This finding indicates that DFQ-ViT can achieve quantization results comparable to or even better than those obtained with real data calibration, without accessing the original training data. This is of great significance for practical application scenarios where the use of original data for quantization calibration is restricted due to privacy or security concerns. It provides an effective solution for deploying Vision Transformer models on resource-constrained devices when data is unavailable.

\begin{table*}[t]
  \centering
\caption{The Top-1 and Top-5 accuracy of various ViT models under different quantization settings and methods after data-free quantization on the ImageNet-1k. Results in \textbf{bold} signify the top performance, and those in \underline{underline} denote the second-best performance.}
\begin{tabular}{c|c|cc|cc|cc|cc|cc}
\toprule    
\multirow{2}[4]{*}{Method} & \multirow{2}[4]{*}{bit-width} & \multicolumn{2}{c|}{DeiT-T} & \multicolumn{2}{c|}{DeiT-S} & \multicolumn{2}{c|}{DeiT-B} & \multicolumn{2}{c|}{Swin-T} & \multicolumn{2}{c}{Swin-S} \\
\cmidrule{3-12}
 &   & Top-1 & Top-5 & Top-1 & Top-5 & Top-1 & Top-5 & Top-1 & Top-5 & Top-1 & Top-5\\
 \midrule
    real data & \multirow{4}[1]{*}{W3/A8} &23.14  &45.69  &44.45  &68.58  &\underline{56.79} &\underline{78.60}  &\textbf{53.43}  &\textbf{76.96}  &54.34  &78.13\\
    gaussian noise  & &3.90  &20.61  &9.09  &42.16  &4.03  &19.36  &0.43  &2.16  &1.73  &9.46\\
    PSAQ-ViT  & &\underline{23.68}  &\underline{46.23}  &\underline{44.76}  &\underline{68.71}  &56.06  &77.99  &\underline{53.40}  &\underline{76.30} &\underline{54.63}  &\underline{78.42} \\
    DFQ-ViT(Ours)  & &\textbf{27.97}  &\textbf{51.53}  &\textbf{45.79}  &\textbf{68.92}  &\textbf{57.50}  &\textbf{79.01}   &52.56  &74.93  &\textbf{57.13}  &\textbf{80.35}\\
    \midrule
    real data & \multirow{4}[1]{*}{W4/A8} &65.38  &86.74  &72.24  &90.72  &76.39  &92.62  &70.29  &89.63  &74.18  &92.12\\
    gaussian noise  & &5.21  &27.65  &16.74  &59.65  &12.39  &44.03  &0.48  &2.34  &4.01  &20.95\\
    PSAQ-ViT  & &\underline{65.56}  &\underline{86.90}  &\underline{72.97}  &\underline{91.27}  &\underline{76.97}  &\underline{92.97}     &\underline{71.66}  &\underline{90.43}  &\underline{75.32}  &\underline{92.73}\\
    DFQ-ViT(Ours)  & &\textbf{65.84}  &\textbf{87.04}  &\textbf{73.69}  &\textbf{91.87}  &\textbf{77.23}  &\textbf{93.02}   &\textbf{74.01}  &\textbf{91.66}  &\textbf{76.24}  &\textbf{92.97}\\
   \midrule
    real data & \multirow{4}[1]{*}{W6/A8} &71.04  &90.42  &\underline{76.19}  &\underline{93.12}  &\underline{78.75}  &\underline{93.98}  &\underline{74.80}  &\underline{92.25}  &76.17  &93.02\\
    gaussian noise  & &9.39  &43.53  &19.92  &66.29  &12.43  &45.25  &0.51  &3.00  &7.92  &33.89\\
    PSAQ-ViT  & &\underline{71.21}  &\textbf{90.60}  &75.63  &92.90  &78.61  &93.86     &74.45  &92.09  &\underline{76.42}  &\underline{93.21}\\
    DFQ-ViT(Ours)  & &\textbf{71.34}  &\underline{90.51}  &\textbf{76.43}  &\textbf{93.17}  &\textbf{79.35}  &\textbf{94.16}   &\textbf{76.61}  &\textbf{93.25}  &\textbf{77.12}  &\textbf{93.37}\\
    \midrule
    real data & \multirow{4}[1]{*}{W8/A8} &71.37  &90.67  &76.37  &93.23  &78.86  &94.00  &74.32  &92.16  &75.47  &92.80\\
    gaussian noise  & &12.28  &51.85  &22.44  &70.10  &20.54  &60.87  &0.56  &3.09  &9.27  &38.89\\
    PSAQ-ViT  & &\underline{71.62}  &\underline{90.78}  &\underline{76.37}  &\underline{93.27}  &\underline{79.20}  &\underline{94.24}     &\underline{75.77}  &\underline{92.94}  &\underline{76.79}  &\underline{93.30}\\
    DFQ-ViT(Ours)  & &\textbf{71.64}  &\textbf{90.84}  &\textbf{77.04}  &\textbf{93.64}  &\textbf{79.51}  &\textbf{94.35}   &\textbf{76.85}  &\textbf{93.49}  &\textbf{77.21}  &\textbf{93.45}\\
    \bottomrule
    \end{tabular}
     
  \label{tab1}
\end{table*}

\begin{figure}[t]
    \centering
    \begin{subfigure}[b]{0.48\textwidth}
        \includegraphics[width=\textwidth]{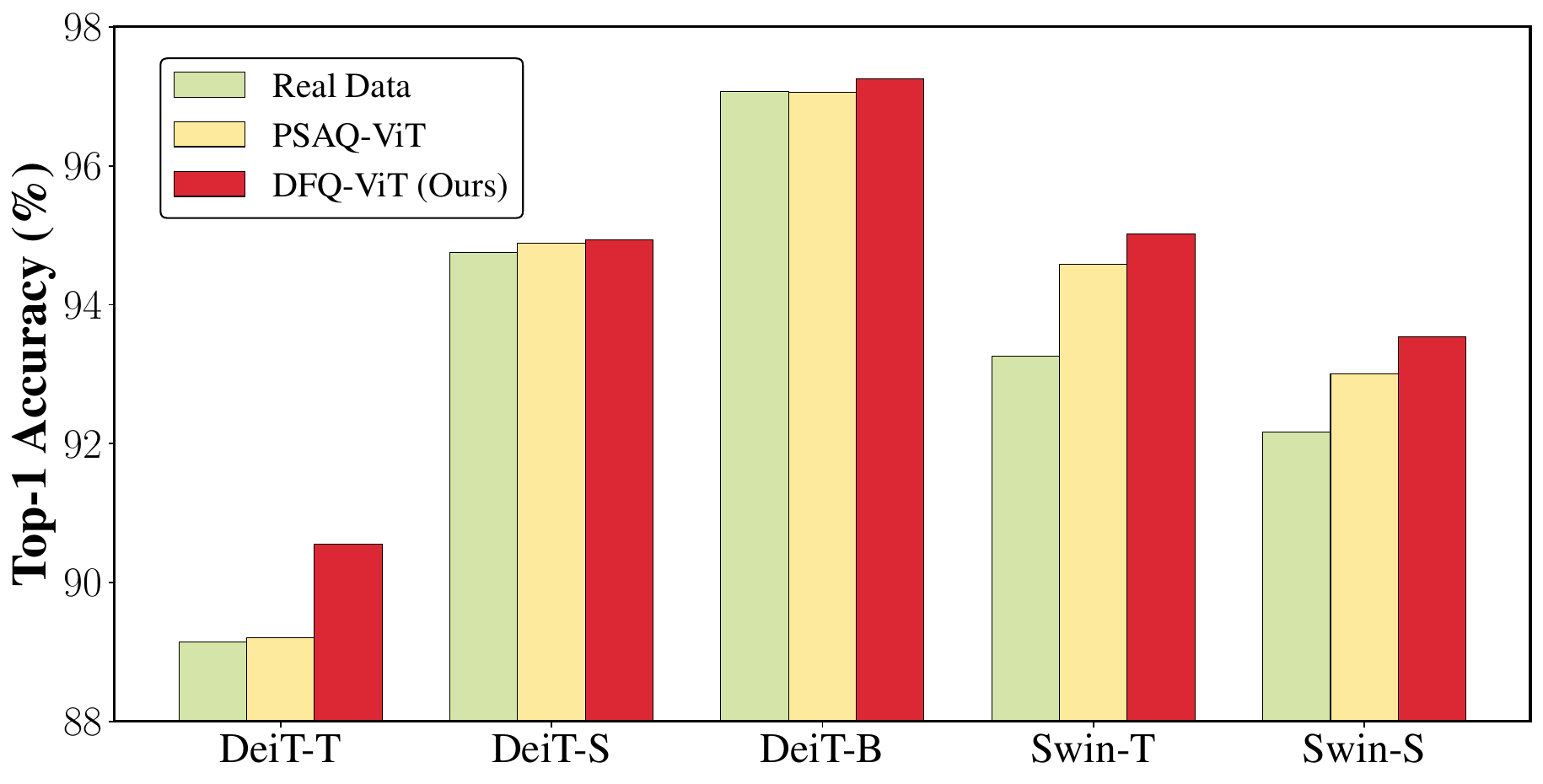}
        \caption{ CIFAR-10 dataset.}
        \label{fig:subfig1}
    \end{subfigure}
    \begin{subfigure}[b]{0.48\textwidth}
        \includegraphics[width=\textwidth]{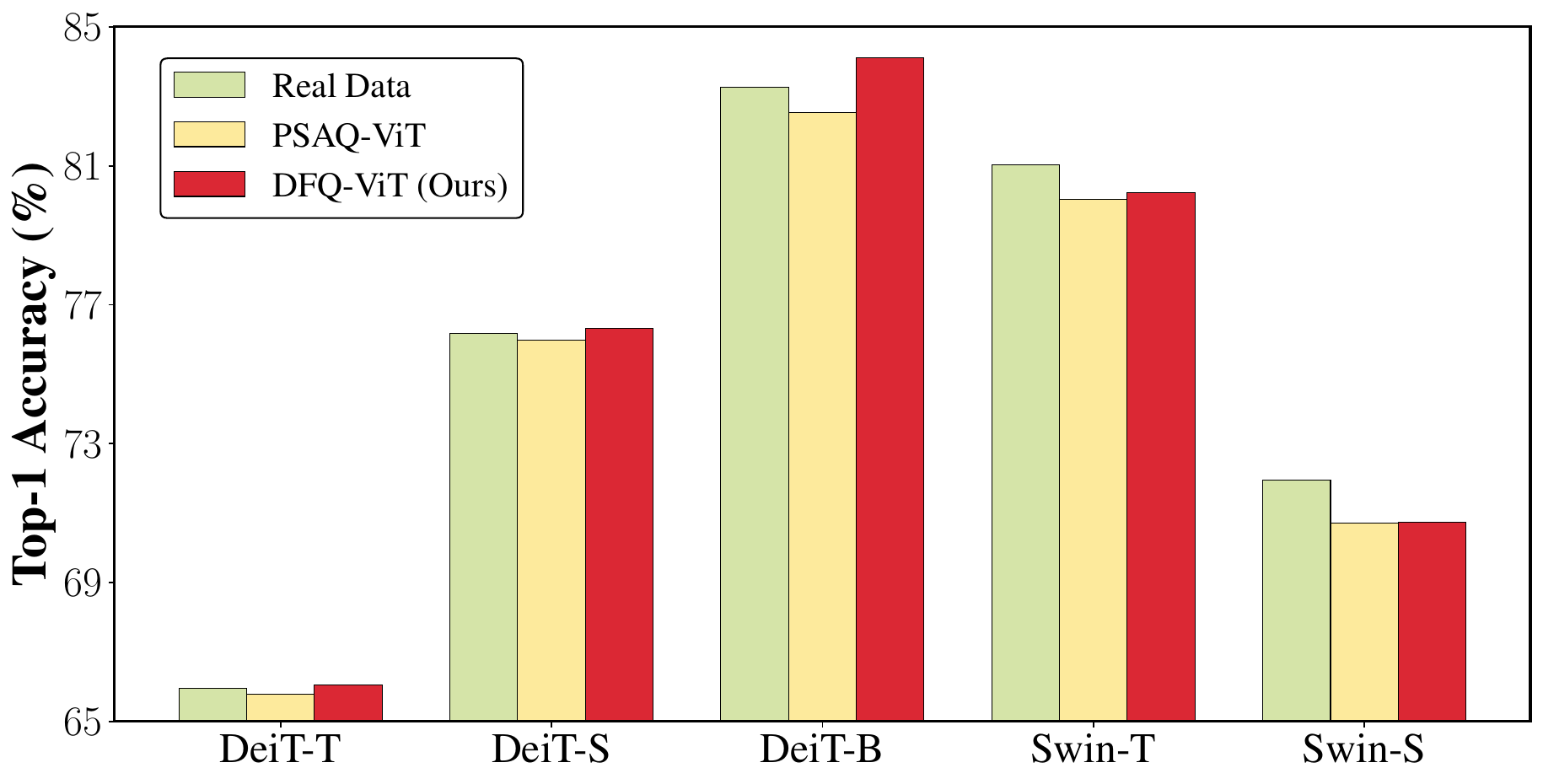}
        \caption{CIFAR-100 dataset.}
        \label{fig:subfig2}
    \end{subfigure}
    \caption{Top-1 accuracy of various ViT models under W4/A8  quantization settings after data-free quantization on the CIFAR-10 and CIFAR-100.}
    \label{cifar}
\end{figure}

Moreover, as shown in Figure \ref{cifar}, we evaluate DFQ-ViT on the CIFAR-10 and CIFAR-100 datasets. Since calibrating the model using Gaussian noise resulted in poor performance, we do not include it in the figure. On the CIFAR-100 dataset with a 4-bit weight and 8-bit activation (W4/A8) quantization setting, models quantized using real data demonstrate excellent performance. Specifically, the Top-1 accuracy of DeiT-T, DeiT-S, DeiT-B, Swin-T, and Swin-S are 65.96\%, 76.17\%, 83.27\%, 81.03\%, and 71.94\% respectively. This indicates that real data can provide an accurate calibration benchmark for quantized models, thereby maintaining high performance. PSAQ-ViT also shows some competitiveness, with the Top-1 accuracy of DeiT-T, DeiT-S, DeiT-B, Swin-T, and Swin-S being 65.78\%, 75.98\%, 82.54\%, 80.03\%, and 70.71\% respectively. In contrast, our proposed DFQ-ViT achieves the best performance on all models, with the Top-1 accuracy of DeiT-T, DeiT-S, DeiT-B, Swin-T, and Swin-S being 66.06\%, 76.31\%, 84.10\%, 80.23\%, and 70.74\% respectively. This demonstrates that the synthetic samples generated by DFQ-ViT can better approximate the distribution of real data, thereby achieving superior quantization performance without relying on original data.

On the CIFAR-10 dataset with the same W4/A8 quantization setting, models quantized using real data also show excellent performance. The Top-1 accuracy of DeiT-T, DeiT-S, DeiT-B, Swin-T, and Swin-S are 89.15\%, 94.75\%, 97.07\%, 93.26\%, and 92.17\% respectively. This further highlights the importance of real data in maintaining the performance of quantized models. PSAQ-ViT performs better than Gaussian noise but still falls short of models calibrated with real data, with the Top-1 accuracy of DeiT-T, DeiT-S, DeiT-B, Swin-T, and Swin-S being 89.21\%, 94.88\%, 97.06\%, 94.58\%, and 93.01\% respectively. In comparison, DFQ-ViT outperforms PSAQ-ViT on all models, achieving a Top-1 accuracy of 90.55\% for DeiT-T, 94.94\% for DeiT-S, 97.25\% for DeiT-B, 95.02\% for Swin-T, and 93.54\% for Swin-S. This indicates that DFQ-ViT excels in generating high-quality synthetic samples that can better represent real data, thereby achieving superior quantization performance on the CIFAR-10 dataset.

\subsection{Ablation Studies}

\begin{table}[t]
\centering
\caption{Ablation studies about varied components of DFQ-ViT on Swin-T. The optimal results are indicated in \textbf{bold}.}
\begin{tabular}{ccccc}
\hline
Method         &bit-width  & Top-1  & bit-width  & Top-1 \\ \hline
Full-precision & W32/A32   & 81.37       & -     & -     \\ \hline
Baseline       &W4/A8    &71.66    &W8/A8     &75.77     \\
+E2H           &W4/A8    &73.26    &W8/A8     &76.62      \\
+ACM           &W4/A8     &72.46    &W8/A8     &76.09      \\
\hline
\textbf{+E2H+ACM}         &W4/A8     & \textbf{74.01}       & W8/A8     & \textbf{76.85}        \\ \hline
\end{tabular}
\label{tab3}
\end{table}

We perform comprehensive ablation studies to demonstrate the efficacy of the proposed components in DFQ-ViT on ImageNet-1k. Table \ref{tab3} demonstrates the impact of the E2H and the ACM on the performance of the quantized models. The results indicate that both E2H and ACM significantly enhance the accuracy of the quantized models when used individually. For instance, at W4/A8 precision, the E2H strategy alone improves the Top-1 accuracy of the Swin-T model from 71.66\% to 73.26\%, while ACM alone achieves a Top-1 accuracy of 72.46\%. When combined, E2H and ACM achieve a synergistic effect, further boosting the Top-1 accuracy to 74.01\%. This suggests that E2H and ACM complement each other by improving the quality of synthetic samples and aligning the intermediate layer activations of the quantized model with those of the full-precision model, respectively. The E2H strategy enhances the quality of synthetic samples by gradually refining the global and local features through a curriculum learning approach. This allows the model to capture more detailed information in the synthetic samples, thereby improving the overall performance. Meanwhile, ACM effectively reduces quantization errors by correcting the intermediate layer activations, ensuring that the quantized model's behavior closely matches that of the full-precision model during inference. The combined use of E2H and ACM thus provides a comprehensive solution to the challenges of data-free quantization for ViTs.

\begin{table*}[t]
  \centering
  \caption{Results with different sample amounts and iteration numbers. When comparing sample amounts, we fix the number of iterations at 500; when comparing iteration counts, the sample amount is set to 16. Results in \textbf{bold} signify the top performance, and those in \underline{underline} denote the second-best performance.}
\begin{tabular}{c|c|cc|cc|cc|cc|cc}
\toprule    
\multirow{2}[4]{*}{Amount} & \multirow{2}[4]{*}{bit-width} & \multicolumn{2}{c|}{DeiT-T} & \multicolumn{2}{c|}{DeiT-S} & \multicolumn{2}{c|}{DeiT-B} & \multicolumn{2}{c|}{Swin-T} & \multicolumn{2}{c}{Swin-S} \\
\cmidrule{3-12}
 &   & Top-1 & Top-5 & Top-1 & Top-5 & Top-1 & Top-5 & Top-1 & Top-5 & Top-1 & Top-5\\
 \midrule
    2 & \multirow{5}[1]{*}{W4/A8} &64.49  &86.15  &71.55  &89.94  &76.33  &92.44  &73.47 &91.15  &75.49  &92.36\\
    4 & &65.00 &86.64  &72.22  &90.64  &76.65  &92.69  &73.62  &91.58  &75.48  &92.54\\
    8  & &65.51  &86.93  &72.43  &90.77  &\textbf{77.50}  &\textbf{93.18}  &\textbf{74.22} & \textbf{91.88} &\underline{76.13}  &\textbf{93.02} \\
    16 & &\textbf{65.84}  &\underline{87.04}  &\textbf{73.69}  &\textbf{91.87}  &\underline{77.23}  &\underline{93.02}   &\underline{74.01}  &\underline{91.66}  &\textbf{76.24} &\underline{92.97}\\
    24 & &\underline{65.79}  &\textbf{87.22}  &\underline{73.16}  &\underline{91.34}  &77.07 &92.93   &73.60  &91.40  &75.44  &92.72\\
    \bottomrule
    \end{tabular}
\begin{tabular}{c|c|cc|cc|cc|cc|cc}
\toprule    
\multirow{2}[4]{*}{Iteration} & \multirow{2}[4]{*}{bit-width} & \multicolumn{2}{c|}{DeiT-T} & \multicolumn{2}{c|}{DeiT-S} & \multicolumn{2}{c|}{DeiT-B} & \multicolumn{2}{c|}{Swin-T} & \multicolumn{2}{c}{Swin-S} \\
\cmidrule{3-12}
 &   & Top-1 & Top-5 & Top-1 & Top-5 & Top-1 & Top-5 & Top-1 & Top-5 & Top-1 & Top-5\\
 \midrule
    100 & \multirow{5}[1]{*}{W4/A8} &65.52 &86.80  &72.56 &91.00  &\underline{77.01}  &\underline{92.95}  &73.31 &91.23  &74.73  &92.34\\
    300 & &65.52 &86.95  &\underline{72.95}  &\underline{91.45} &77.00  &92.94  &73.36  &91.25  &75.72  &92.88\\
    500 &  &\textbf{65.84} &\underline{87.04}  &\textbf{73.69}  &\textbf{91.87}  &\textbf{77.23}  &\textbf{93.02}  &\textbf{74.01}  &\textbf{91.66} &\textbf{76.24}  &\underline{92.97} \\
    600 & &\underline{65.75}  &\textbf{87.05} &72.93 &91.17  &76.94  &92.77   &\underline{73.44}  &\underline{91.32}  &\underline{76.07} &\textbf{93.07}\\
    1000 & &65.66 &86.90 &72.89  &91.06  &76.92 & 92.82  &73.14  &91.20  &75.10  &92.60\\
    \bottomrule
    \end{tabular}
 \label{tab2}
\end{table*}

Table \ref{tab2} explores the influence of the number of calibration samples and the number of iterations per sample on the performance of DFQ-ViT. The results show that the optimal number of calibration samples is 16, as fewer samples (e.g., 4) fail to provide sufficient information for accurate quantization parameter adjustment, while more samples (e.g., 24) introduce noise and outliers that degrade performance. Similarly, the optimal number of iterations per sample is found to be between 500 and 600. Fewer iterations (e.g., 100) result in under-optimized synthetic samples, whereas excessive iterations (e.g., 1000) may lead to overfitting and reduced generalization. These findings highlight the importance of balancing the number of samples and iterations to achieve the best performance in data-free quantization. Specifically, using 16 calibration samples and 500 iterations per sample provides the best trade-off between computational efficiency and model accuracy. This optimal setting ensures that the synthetic samples are of high quality and that the quantization parameters are well-calibrated, leading to improved performance of the quantized model. The results further demonstrate the robustness and adaptability of DFQ-ViT under different calibration conditions, confirming its effectiveness as a data-free quantization method for Vision Transformers.

\subsection{Synthetic Samples Analysis}
\begin{figure*}[t]
\centering
\includegraphics[width=1\textwidth]{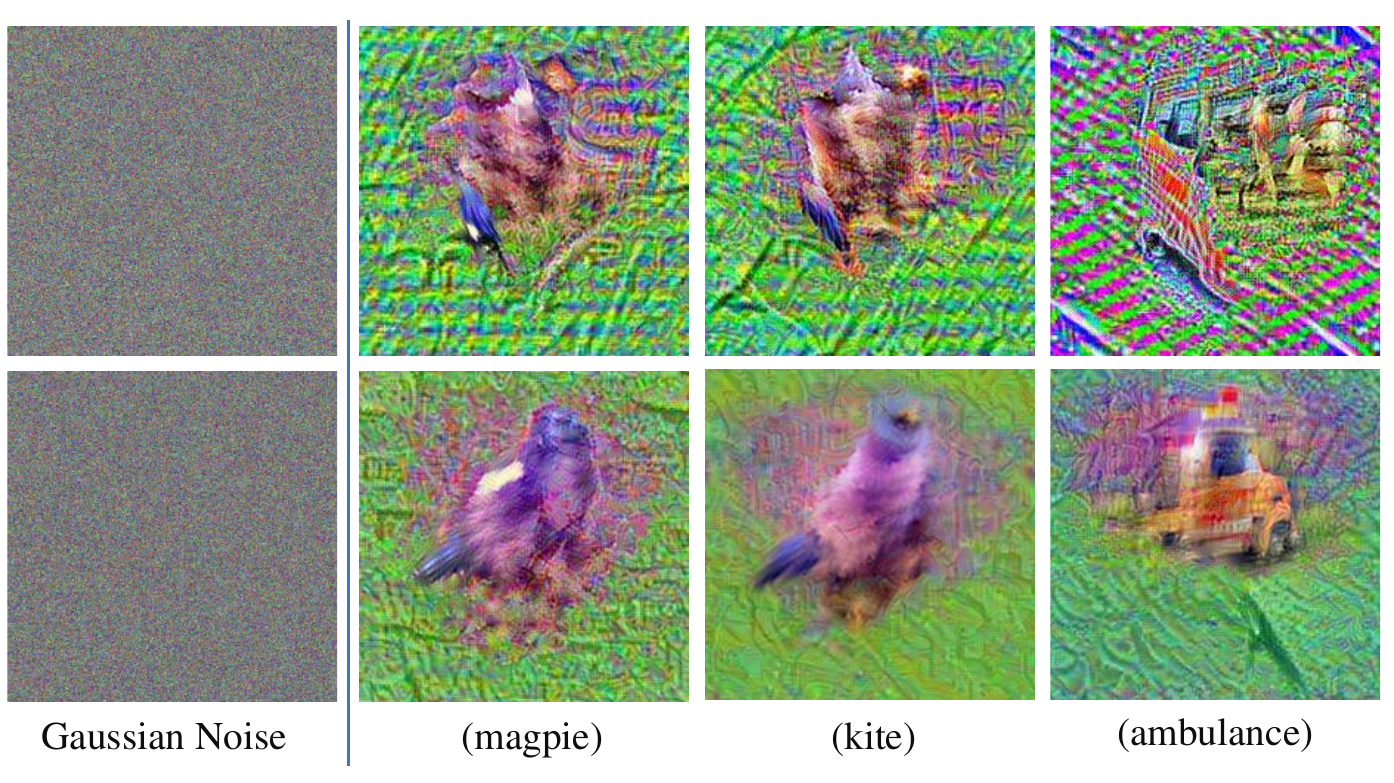} 
\caption{Visualization of synthetic samples: The first column is Gaussian Noise. For columns from the second onward, the first row is generated by PSAQ-ViT, and the second row by DFQ-ViT. The second row shows significantly better quality due to DFQ-ViT's superior balance of global and local information.}
\label{fig3}
\end{figure*}
Figure \ref{fig3} shows the visualization results of synthetic samples generated using Gaussian noise, PSAQ-ViT, and DFQ-ViT. The samples generated by DFQ-ViT exhibit significantly improved quality compared to those produced by existing methods, primarily due to the E2H strategy employed by DFQ-ViT. In the initial stage of sample synthesis, the E2H strategy uses larger cropping areas to capture the global structure and outline of objects. As the synthesis progresses, the size of the cropping area gradually decreases, allowing the model to focus on refining local details. This step-by-step refinement from global to local features enables the synthetic samples to better simulate the complexity and characteristics of real data. For example, in the “kite” category, the synthetic samples generated by DFQ-ViT clearly display details such as the tail feathers of the kite, which are almost unobservable in the samples produced by PSAQ-ViT. This ability to balance global and local information significantly enhances the overall quality of the synthetic samples, making them more suitable for calibrating the quantized model.

The visualization in Figure \ref{fig3} also highlights the limitations of existing methods. Synthetic samples generated by Gaussian noise lack both global structure and local details, resulting in poor quality that is insufficient for accurate model calibration. Although PSAQ-ViT represents a significant improvement over Gaussian noise, it still struggles to capture fine local features due to its neglect of the order and difficulty of sample synthesis. In contrast, the E2H strategy of DFQ-ViT effectively addresses these issues by progressively optimizing the samples from simpler to more complex tasks. This approach not only improves the quality of the synthetic samples but also enhances the calibration accuracy of the quantized model, ultimately leading to better performance in the inference stage.

\subsection{Visualization of Intermediate Synthesis Processes}
\begin{figure*}[t]
\centering
\includegraphics[width=1\textwidth]{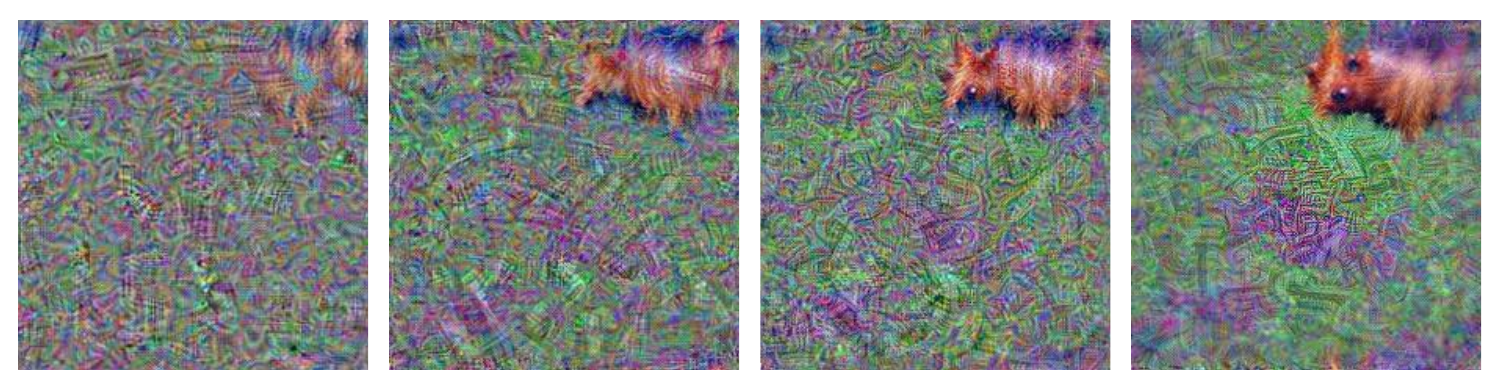} 
\caption{Visualizations of { 100, 200, 300, 500} steps
(salamander)}
\label{fig4}
\end{figure*}

Figure \ref{fig4} presents the visualizations of synthesized images for the "salamander" category at different iteration steps (100, 200, 300, and 500 steps). By examining these images, we can gain insights into how the synthesis process gradually refines and optimizes the synthetic samples. In the early stages of iteration (100 steps), the synthesized images appear blurry, with only the general outline and basic shape features of the target category being captured. As the iteration progresses (200 steps), the contours of the images become clearer, and some basic textures and local features begin to emerge. This indicates that the synthesis process is learning to represent the key features of the target category more effectively. By 300 steps, the synthesized images show richer details and more refined textures, with the overall visual appearance becoming closer to that of real images. This demonstrates that, with more iterations, the model is better able to balance the generation of global structures and local details, thereby improving the quality of the synthesized samples. Ultimately, at 500 steps, the synthesized images exhibit a highly realistic appearance, with both complete global structures and well-represented local details such as textures and colors. This underscores the efficiency and effectiveness of the synthesis process, which can produce high-quality synthetic samples once a sufficient number of iterations have been completed. These samples can then serve as valuable inputs for calibrating quantized models.
Through this series of visualizations of intermediate synthesis processes, we can clearly observe how the synthesis process transitions from rough outlines to high-quality images with rich details. This gradual refinement not only enhances the quality of the synthesized samples but also provides an intuitive understanding of the model's generation mechanism. Moreover, it validates the effectiveness of the adopted synthesis strategy in generating representative and high-quality synthetic samples, which is crucial for the successful implementation of data-free quantization methods.

\subsection{Parameter Quantity Analysis}
Table \ref{tab4} displays the original parameter counts of various ViT models and their corresponding ACM parameter counts. It can be observed that the parameter count of ACM is negligible compared to the millions or even tens of millions of parameters in the models. Specifically, DeiT-T has a parameter count of 5.72M, and its corresponding ACM parameter count is only 1192, accounting for 0.020\% of the original model’s parameters. The parameter count of DeiT-S is 22.05M, and the corresponding ACM parameter count is 1384, which is 0.006\% of the original model’s parameters. For the larger DeiT-B model with 86.57M parameters, the ACM parameter count is 1768, representing only 0.002\% of the original model’s parameters. Similarly, Swin-T has 28.29M parameters with an ACM parameter count of 1768 (0.006\%), and Swin-S has 49.61M parameters with the same ACM parameter count of 1768 (0.004\%). Therefore, ACM essentially does not increase computational overhead, making it a highly efficient approach to improve the performance of quantized models without adding significant complexity.

\begin{table}[htpb]
\centering
\caption{Original model and ACM Parameter Quantity Analysis. We compare the number of parameters in the original model with those in the ACM, where the unit of the number of parameters in the original model is millions (M).}
\begin{tabular}{c|ccc}
\toprule 
Model         &Model params(M)   &ACM params   & Ratio   \\ \midrule 
DeiT-T &5.72   &1192      &0.020\%        \\ 
DeiT-S &22.05    &1384    &0.006\%         \\
Deit-B &86.57    &1768    &0.002\%         \\
Swin-T &28.29     &1768    &0.006\%           \\
Swin-S &49.61     &1768       &0.004\%       \\  \bottomrule
\end{tabular}
\label{tab4}
\end{table}

The negligible parameter count of ACM is a crucial advantage for practical deployment, especially on resource-constrained devices. Since ACM introduces minimal additional parameters, it ensures that the overall model size and computational requirements remain almost unchanged. This is particularly important for applications where memory and processing power are limited, such as edge devices or mobile platforms. The small parameter overhead of ACM allows DFQ-ViT to be effectively deployed without compromising the benefits of quantization in terms of reduced memory usage and faster inference speed.

\subsection{Efficiency Analysis}
We conduct a detailed efficiency analysis to evaluate the computational overhead introduced by the proposed DFQ-ViT method compared to the baseline PSAQ-ViT. As shown in Table \ref{tab5}, both methods spend the majority of their time in the synthetic sample generation phase, which is a critical step for calibrating the quantized models without access to real data. Specifically, DFQ-ViT and PSAQ-ViT both require 125 seconds to generate synthetic samples for the DeiT-T model. This indicates that the E2H strategy employed by DFQ-ViT does not introduce additional computational cost during sample generation, maintaining the same efficiency as PSAQ-ViT. In the calibration phase, DFQ-ViT takes slightly longer (0.093 seconds) compared to PSAQ-ViT (0.076 seconds), primarily due to the computation of the ACM. Nevertheless, this increase is relatively minor and does not have a substantial effect on the overall efficiency. During the inference phase, both methods achieve the same inference time of 0.016 seconds per image. This demonstrates that the additional matrix operations introduced by the ACM in DFQ-ViT do not add noticeable computational overhead during inference, ensuring that the performance gains achieved through DFQ-ViT come without sacrificing inference speed.

\begin{table}[htpb]
\centering
\caption{Efficiency analysis of PSAQ-ViT and DFQ-ViT on DeiT-T. The number of synthetic samples used for calibration is 16, and the number of iterations is 500.}
\begin{tabular}{cccc}
\hline
Method         &Generation(s)  & Calibration(s)  &Inference(s)   \\ \hline
PSAQ-ViT          &125    &0.076    &0.016          \\
DFQ-ViT           &125     &0.093    &0.016           \\
\hline
\end{tabular}
\label{tab5}
\end{table}

The efficiency analysis highlights the practical applicability of DFQ-ViT in real-world scenarios. The negligible increase in calibration time and the unchanged inference speed make DFQ-ViT a highly efficient solution for data-free quantization of Vision Transformers. This is especially crucial for deploying on devices with limited resources, where computational power is restricted and inference speed is of utmost critical. By maintaining the same level of efficiency as PSAQ-ViT while significantly improving the quantization performance, DFQ-ViT offers a superior trade-off between accuracy and computational cost. 

\section{Conclusions}
This research presents a novel data-free quantization approach for ViTs, termed DFQ-ViT, which focuses on enabling the quantization of ViTs in environments where original data cannot be accessed and computational resources are restricted. By introducing an Easy to Hard  sample synthesis strategy and an Activation Correction Matrix, DFQ-ViT achieves significant improvements in enhancing the quality of synthetic samples and aligning the intermediate layer activations of the quantized model. During the sample synthesis phase, the E2H strategy gradually adjusts the cropping ratio from large to small, enabling the synthetic samples to learn from global structures to local details step by step. This approach better simulates the complexity of real data. In the model calibration phase, the ACM adjusts the activations of the intermediate layers in the quantized model to make them closer to the activation patterns of the full-precision model, thereby  reducing the  quantization errors. Through optimizing the quality of synthetic samples and rectifying the activations in intermediate layers, DFQ-ViT delivers an efficient and effective method for quantizing Vision Transformers in settings where original data is unavailable and resources are constrained.

\bibliographystyle{ACM-Reference-Format}
\bibliography{main}

\end{document}